# New Ideas for Brain Modelling 7

Kieran Greer, Distributed Computing Systems, Belfast, UK.
http://distributedcomputingsystems.co.uk
Version 1.1

*Abstract*—This paper updates the cognitive model, firstly by creating two systems and then unifying them over the same structure. It represents information at the semantic level only, where labelled patterns are aggregated into a 'type-set-match' form. It is described that the aggregations can be used to match across regions with potentially different functionality and therefore give the structure a required amount of flexibility. The theory is that if the model stores information which can be transposed in consistent ways, then that will result in knowledge and some level of intelligence. As part of the design, patterns have to become distinct and that is realised by unique paths through shared aggregated structures. An ensemble-hierarchy relation also helps to define uniqueness through local feedback that may even be an action potential. The earlier models are still consistent in terms of their proposed functionality, but some of the architecture boundaries have been moved to match them up more closely. After pattern optimisation and tree-like aggregations, the two main models differ only in their upper, more intelligent level. One provides a propositional logic for mutually inclusive or exclusive pattern groups and sequences, while the other provides a behaviour script that is constructed from node types. It can be seen that these two views are complimentary and would allow some control over behaviours, as well as memories, that might get selected.

## 1  Introduction

This paper updates the cognitive model, firstly by creating two systems and then unifying them over the same structure. It represents information at the semantic level only, where labelled patterns are aggregated into a 'type-set-match' form. It is described that the aggregations can be used to match across regions with potentially different functionality and therefore give the structure a required amount of flexibility. The theory is that if the model stores information which can be transposed in consistent ways, then that will result in knowledge and some level of intelligence. As part of the design, patterns have to become distinct and that is realised by unique paths through shared aggregated structures. An ensemble-hierarchy relation also helps to define uniqueness through local feedback that may even be an action potential. The earlier models are still consistent in terms of their proposed functionality, but some of the architecture boundaries have been moved to match them up more closely. After pattern optimisation and tree-like aggregations, the two main models differ only in their upper, more intelligent level.

The behaviour model [7] gives a geometric progression, with memory structures at the base moving from nested ensembles to unique tree sets. There is then an interface with the upper experience-based level that models the behaviours, with a scheduling layer at the very top. The original cognitive model can use a similar memory structure, but its upper level has been changed to a propositional logic of mutually inclusive or exclusive pattern groups and sequences. The logic can complement the behaviour actions, to provide some level of control over their execution and it in fact makes use of the layers and structures of the original model, only with slightly different architectural boundaries. This paper proposes that the memory structures can be formed in a largely unsupervised manner, where they represent and organise data using limited sensory processes. The patterns can then be corrected through interactions with the experience-based levels [4]. Key to this is a short feedback and reinforcement loop between the ensemble and the hierarchy. Rote learning is therefore still part of the system, but 'units of work' are created first through self-organisation, before they can be further clustered into knowledge trees, for example, using more direct approaches. The design also recognises that transitions from one type of information to the next are required and different types of transposition are described. As such, knowledge and experience can be merged and interact between brain regions.

The rest of the paper is organised as follows: section 2 lists some related work. Section 3 then describes a 'unit of work' structure, which is more complex than a single neuron and can help to aggregate ensemble patterns. Section 4 gives details about the different types of knowledge and information transpositions that would make the network sufficiently flexible. Section 5 then re-visits the 3-level architecture and compares the two detailed cognitive models, to show that they are essentially the same. Finally, section 6 gives some conclusions on the work.

## 2  Related Work

Because this paper is a consolidation of earlier work, any of the author's papers [4]-[16] are of interest. Other research is also mentioned, to show that a lot of the established theories are involved. As the network is being described as semantic, Semantic Networks [24] are obviously of interest and there is a return of Stigmergy [5][3] as a fundamental part of the organisation. The models in this paper are more for the computer program and so the supporting biology theory is not as important, but global properties from statistical physics [26], for example, is still very important. One other biological aspect is the Neural Binding Problem [4] that asks how distinct concepts can be understood, still as distinct concepts, when they are combined. For example, 'why don't we confuse a red circle and a blue square with a blue circle and a red square'. This problem is well-known but poses problems when trying to understand it in terms of the human brain. For a semantic computer model however, it is relatively easy to find a solution, as described in section 4 and later. That paper also quotes research in [2], which shows that coupling between activity in distant brain areas may be mediated by local field potentials (LFP) and phase coupling.  This is interesting as part of the ensemble-hierarchy structure [10][11], discussed in section 3 and later. With regard to a broad framework, and-or graphs, along with theorem-proving are written about in [20]. Consistent with the earlier papers, the and-or graphs would still be constructed between more vertical tree structures (And) and the more horizontal symbolic neural network (Or) [16]. The clustering in this paper would be part of the 'upwards' memory formation. The theorem-proving trees of axioms to goals can be placed in the top 'intelligent' level and also be part of the 'downwards' search, matching with the memory structures. The paper [25] also models behaviours and it is interesting that they consider the behaviours to be unique (time or sequence-based) sets of event patterns that are then clustered, rather than existing as each individual event. The idea is to mix and match over unique pattern sets.

Because the architecture is neural, quite a lot of the neural network theory could become relevant and replace any of the author's own clustering algorithms. One of the earliest papers [21] modelled the neural linking very closely and even suggested logic rules. They argued that if there is a definite set of rules or theorems under which the neurons behave, this can then be described by propositional logic. They then focused more on behaviour and states, rather than precise mathematical values, but this is also the case for the author's model and one of the detailed designs [12] has been converted to represent some level of propositional logic instead. Stability in the model is important, where the real brain has developed mechanisms to correct itself and realise stable states. For example, there is evidence that brain activity during sleep employs a Boltzmann-like (or Hopfield) [23] learning algorithm, in order to integrate new information and memories into its structure [27]. The middle layer of the cognitive model might benefit from a Boltzmann-like machine, although a different self-organising unit is currently suggested. The idea of a unit would give the patterns some shape to achieve a stable state with, and stability also means a lower-energy state, which is always important.

## 3   The Neural Unit

It has been useful to look at a neural assembly as a 'unit of work' instead of single neurons. This was suggested at the beginning with the ReN (refined neuron) [12] and then later with a subsequent ensemble-hierarchy structure [10][11]. The intention of the ReN was to make the neural signals more analogue by aggregating a set of input neurons through a single output neuron, where each input neuron then contributes only a fraction of the output signal. The intention of the ensemble-hierarchy structure was to make the signal it produces more distinct, where the hierarchy provides structure and would fire with the ensemble, giving it that extra definition, rather like an action potential. The argument being that a fully-connected ensemble pattern is easier to find and activate in the first place. It may be too fanciful to imagine that there is an ensemble-hierarchy structure that will produce these unique 'notes' to define a pattern, but the fact that structure is added to the unit gives it merit and a difference in action potential has been shown to influence activity in a real brain [2].

The ensemble-hierarchy was originally designed with a diverging Concept Tree [14][9] hierarchy part. In fact, the second Concept Tree paper [9] was changed over some versions, where there was confusion as to whether the structure would diverge from the top to the bottom or the bottom to the top. The search process can also be in either direction and so it may be preferable to give priority to finding the concept as part of the search. It is possible to make the argument very organic and write about signal strengths and wiring lengths, but organising neurons considers signal types and chemical compounds as well and so the argument for this paper is purely functional. It shall be kept as to what each layer should do and not exactly how it may be formed. The self-organising unit is shown in Figure 1 and described in the next sections.

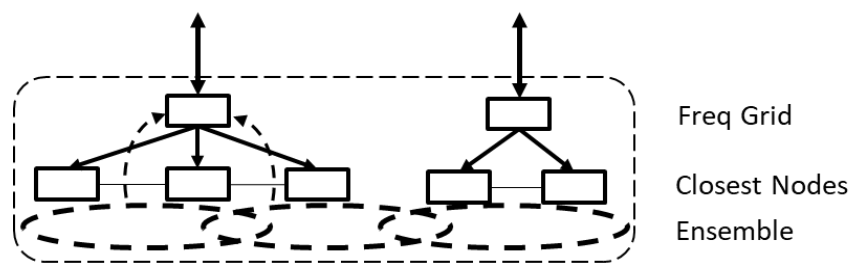

Figure 1. Self-organising 'Unit of Work'.

### 3.1 Self-Organising Unit

The paper [4] has suggested a self-organising algorithm that uses brain-inspired full-linking [1][18] plus a new Frequency Grid [10] count, which produces an ensemble at the base and a hierarchy above that, where the hierarchy is for agglomerative clustering. The frequency grid may also be compatible with a brain architecture. Considering doctrines similar to Hebb's rule [19] 'neurons that fire together wire together', or 'equal distanced neurons fire together', then this is approximately what the frequency grid does. It produces an association count for every node with every other node and the largest count with some other node defines that node's relation. The node clusters are created from events. Consider therefore, that the self-organised units form fully-connected patterns that then fire together, more often than with other nodes. It is also interesting that adding more frequency grid levels produces a much more regular frequency count inside of each unit. The counts in the first level are much more variable and so this could suggest that the nodes become more synchronized with the unit cluster they belong to, in further levels. That would probably encourage Hebbian growth in the mini-cluster only. These units that aggregate individual nodes can then be aggregated themselves in a third level, when we get the self-organising algorithm described in [4]. The algorithm produces something that is thus also like a unit of work.

### 3.2   Ensemble-Tree Hierarchy

It might be easier to invert the tree structure in the ensemble-hierarchy, where the aggregated base node is still added first, but also at the furthest distance and then the sub-concept nodes are added as branches towards the ensemble itself. When the ensemble first fires, an aggregated node is realised some distance away. Possibly a slower firing rate can give more distance. Then if ensemble parts fire, they realise other aggregated nodes in-between. Possibly a faster firing rate encourages the node to form more quickly and therefore also form closer. The Concept Tree counting rule is therefore maintained, but inverted, because the root node, representing the whole ensemble, fires the least and each individual part fires more often, as a single unit. If that is the case, then both more object neurons and also descriptor ones can be closer to the sensorised ensemble. Also for economy of energy, it might be the case that the tree structure should converge and grow backwards, as shown in Figure 1. Each aggregated tree node can still fire somewhere else and represent that part of the ensemble. The structure can probably be looked at as a bit columnar, with a partially-defined outside edge and it is limited by the ensemble base. The design would still like to have secondary links between each neuron and its sibling pattern in the ensemble.

Consider that the ensemble input would be general sensory input from the human body. The tree may also receive input from other regions, which could be the more intelligent cortex areas, for example. If the tree does not receive input from anywhere else, there is no reason for it not to fire in the same way as the ensemble and so the aggregated tree nodes can be a type of matching between different regions, maybe a transposition area. The same process for ensemble-to-ensemble does not ring quite as true. For one thing, aggregated nodes cannot be nested and so they would have to be linked trees and that may force some level of order. Then what is the interface between the two ensembles, it must simply be synchronized firing, but the nesting would cause confusion.

## 4    Knowledge Transitions

The network is made up of knowledge-based and experience-based layers that transpose. Experience is more unstructured and requires time-based events to give it meaning. It transposes into knowledge and uses the time-based events to create the structure. Time will simply allow more unrelated structure through the firing sequences. It may be helpful to think that knowledge-based patterns fire inwards and experience-based patterns fire outwards. This also means that knowledge patterns can be fully-connected and would make suitable terminal states. Knowledge-to-knowledge is also possible and can be a change in object associations, also influenced by time. Because knowledge has structure, it will fire more concepts vertically when active, which gives rise to a type of ensemble in the next layer and so the next layer should probably be shallower, or more horizontal. It is also proposed that transitions from one type of knowledge to the next is what makes the network flexible.

To start with, the problem of shallow or deep hierarchies can be re-visited, with the idea of trying to define when the system might use either in terms of the functionality. There are probably 3 types of information transition:

1. Experience-to-Knowledge.
2. Knowledge-to-Knowledge.
3. Knowledge-to-Experience.

It is proposed that the first transition type produces deeper network structures and probably produces 'a-priori' knowledge. Experience-based information is about the use of objects and so it is more concerned with interactions between them and would thus probably be represented by shallow hierarchies with a lot of lateral links. This can be seen in the top 'intelligent' level of any of the previous the cognitive models [15][12][7]. If the experience is being transitioned into knowledge, then the chances are that it is being converted into a more static and permanent form. The links would be more through the separate parts of each individual object, than across the objects. This is also generally accepted when learning something. It requires the practical experience to learn the tacit knowledge that is not easily defined for the individual person. These extra bits of information that are not obvious would make the network structure deeper and the experience would want to make a permanent representation of it - not of more experience, but of the results of it in the real world. The second type would probably be a transition from a deeper network to a shallower one and is again explicit and a-priori. The static objects in the first knowledge level would be aggregated or compared in the second level, also with the idea of cross-referencing them. It would not make sense to try to add deeper knowledge again - that would happen in the first layer and so the second layer would be about the object interactions instead. The third type of transition is maybe slightly different in that the links produced might be peculiar to the experience type itself. This probably produces implicit and 'a-posteriori' knowledge. For example, induction reasoning would be used to gain new insights from what is already known about. Experience-to-Experience looks unlikely. Why would this occur without some permanent learning first?

The two hierarchy types can also retain their original functionality. The deeper hierarchy would be explicit, but also where the tacit knowledge is kept and understood. The shallow hierarchies would like to deal with whole concepts, such as 'square' or 'circle' and solve the neural binding problem [4], for example. For a computer model, a shallow hierarchy of only two levels or so can solve it through links from leaf nodes in one tree to a similar base node in another tree, as shown in Figure 2. This knowledge problem may be declarative in nature (knowledge of something), but it is not yet events or actions, which is procedural knowledge (how to do something). Procedural knowledge can be stored as pattern sets that represent actions instead of images, for example. To solve the neural binding problem requires distinct representations for each concept. Then simply through the associations that occur, paths of links will form that can define each concept set uniquely. A problem with the tree structure may be that the number of nodes required, makes this impractical. Everything needs to be repeated for everything else, and so a better solution, suggested in section 5.1, is to repeat the list of nodes in two separate layers instead.

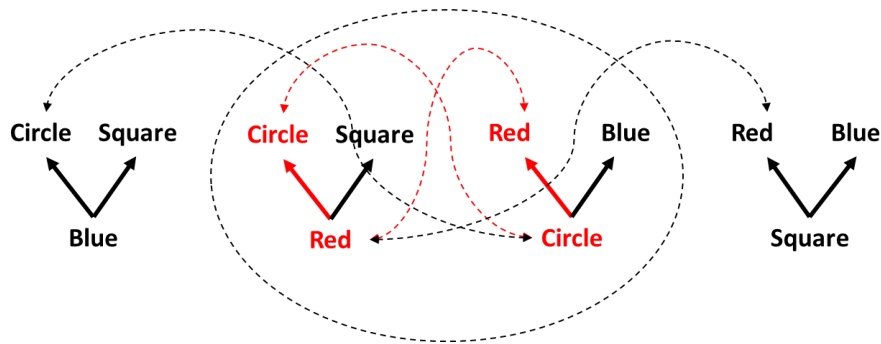

Figure 2. One level of linking in a tree-like temporal model defines a particular ensemble mix [11], figure 2.

### 4.1    Neural Units can produce Unique Representations

The neural units described in section 3 are converging. It would also be helpful if these units can be made unique in their concept representation and that could be quite straightforward through a mix and match approach. If you always walk down the same street, for example, then you will tend to see the same shops and people, but maybe there are different combinations of these each time. At the end of the day, this might be thought over and different scenarios that are related, sorted as clusters. It may be like having a large dataset of values stored in memory and learning different subsets of it each time. The architecture for these unique sets can be seen in the bottom half of Figure 6 or Figure 3, for example. Each ensemble cluster has converged into a much smaller set of aggregated category nodes, possibly using the self-organising units to create the structure. Because the base ensemble does not have the same structure, the input stimulus can activate shared ensemble pattern sets and so there is overlap in the ensemble space.

## 4.2     Ensembles to Types

This is a key requirement for making the network flexible and to give it the ability to discriminate over the pattern structures. In fact, it turns out that we already have the solution to this problem and it is simply properly constructed semantic networks [24], although they cannot solve everything, such as understanding text or mathematics, for example. Consider a nested ensemble of patterns that form essentially as an imprint of the input senses. While the sense might still be used during recall, how is the pattern processed when the input sense is removed? This is also a question about optimisation and what the optimisation is for. The optimisation is to isolate one pattern from other ones, so that its firing event is clearly defined, leading to two choices:

1. Each input pattern is separate from every other one and it fires inwards, to recognise its own structure and no other structure.
2. Each pattern is reinforced from some other level, to boost the features in it and therefore help them to be recognised.

If option 1 is correct, then there is a greater limit on how many patterns and therefore memory images can be stored. Option 2 looks more likely and it can make use of information transitions. And when we use existing knowledge to solve a new problem, we must surely be using option 2. To make a complete process, at least 3 levels of recognition are required. We have the nested ensemble at the base that might be an imprint of the input senses. A level above this is a set of features, represented by smaller aggregated node sets. The aggregated sets might even be single nodes for each feature pattern and can therefore be stored separately more easily, while the nested ensembles share. Each feature set also represents an object in the real world and so it has a further link to another single neuron, for example, that is the label of that object type. Therefore, from a type, we go to feature sets of the type and then to matching patterns at the base. This looks a bit like Figure 1, but each unit would then also link to a type node, not shown.

There is also the possibility of a local feedback between the ensemble and the feature set, where if one fires, it reinforces the other and encourages that region to fire over other regions. Consider for example, a boat or a house. Both have windows and doors, but a house also has walls and a roof, while a boat has a hull and a deck, for example. The window feature is recognised as a window type in the third level and that level links down to both the boat and house ensembles but not their types. The house feature set also has wall and roof features that would have to be active for the type to fire, while the boat set with the hull and sail would not fire as strongly for that set of features. Therefore, the house ensemble is reinforced more and the window gets associated with the house and not the boat. With this network, we get a 'type-set-match' structure, illustrated in Figure 3.

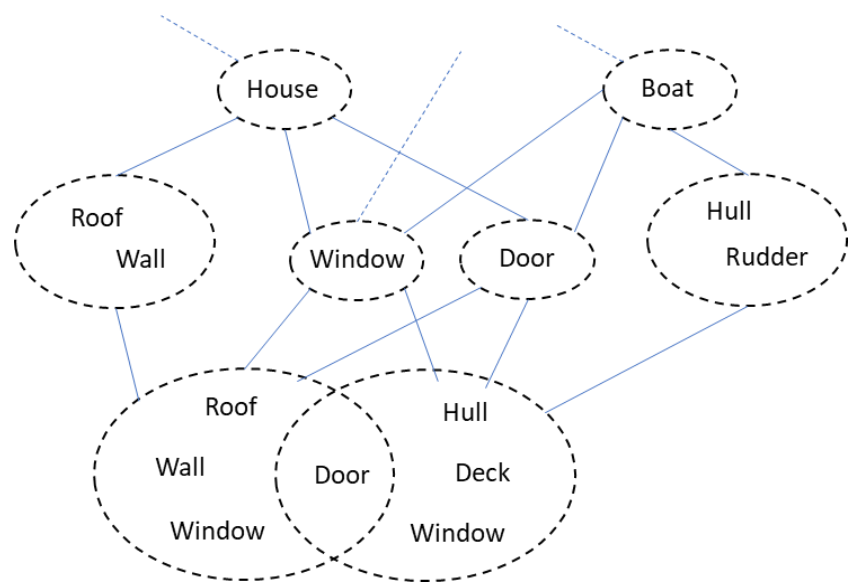

Figure 3. Semantic network with a 'type-set-match' structure. Types of Window, Door, House and Boat are shown.

It might be interesting to think that if two input ensembles had some similarity, then they might naturally form closer to each other. With the boat and the house, for example, the doors are essentially the same and so if the memory for a house door already existed, along with maybe other elements, then if the boat stimulus is input, it might match with some of the existing house pattern and be encouraged to form in that region. This would also be a natural order and indexing system, because when the human goes to think about houses, he/she may think next about boats and not about something that is completely unrelated.

Stigmergy [5][3] may play a part here, because the imprint of the shared features in the region encourages the rest of the pattern to form there. It would also be the most economic in terms of energy [13].

## 5    The Cognitive Models

Details of two cognitive models have been described in [12] and [7] and are listed in Appendix A. The cognitive model has evolved since the first version [15], which was a lower optimising layer, followed by a middle aggregating layer and then an upper cognitive layer. While the model has been refined in places, it retains the idea of memory patterns followed by aggregated patterns, followed by more intelligent structures and that will have to be the template that is used. The difference in the models is in fact the upper cognitive layer. The first detailed model, described originally in [12] is shown in Figure 5. It used simply single links between sets of complex concepts, but it was decided that these single links did not provide enough information to allow the concepts to trigger each other usefully. It has since been decided that the more intelligent layer could provide some form of controlling logic instead, inspired by the original McCulloch and Pitts paper [21]. A second model was then developed [8][7] with a novel upper layer, shown in Figure 6, that would control behaviours instead. The two detailed designs should also carry equal importance. While the figures look just like sets of concepts with links, it is important to remember that there is an underlying mathematical formula that can construct this in a consistent manner. The transposition from ensemble domains (deep trees) to chunks of knowledge (shallow trees) looks to be consistent. The chunks can be hierarchical, where some tree parts are shared, but the full path through them is unique. The chunking therefore creates sets of knowledge that have unique consequences/actions/results. Sharing the information is not just about economy, but also about search, where if one sub-part is found, it can link back to all of the different possibilities that make use of it.

### 5.1    The Procedural Logic Model

Figure 4 illustrates the new upper layer for the procedural logic model and it can be compared with the square types-level and upwards in Figure 6. In Figure 4, the oval nodes are the aggregated sets and paths through them should be unique. They might initially link with every concept in the bottom logic layer, which is a different type of knowledge and therefore an interface, represented by the square nodes and similar to Figure 2. In this figure however, the tree structure with repeated node sets as branches has been replaced simply with two levels of the same nodes. This is much more economic and may make the structure practical. Each node then links to itself between the layers and also to the other concepts it is related with. Having so many potential links between the two levels can lead to many circuits firing at the same time. Inhibitors would therefore be the main mechanism to control the firing sequences. For example, both avoiding or speaking to a person is possible and they can both fire at the same time as part of the person circuit. There would therefore be an inhibitor between the two sets, so that one of them would switch off.

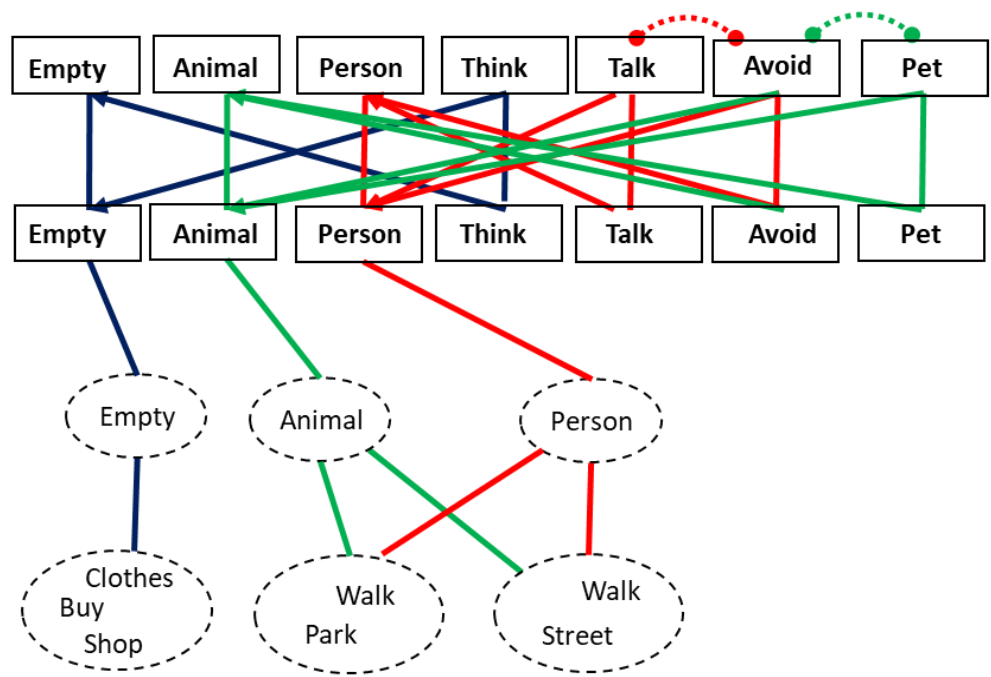

Figure 4. Procedural Logic, based on 2 similar layers that join.

The 2 layers would match with the bottom layer of the symbolic neural network in the original design (Figure 5) and also time-based events. Forming constructive sequences looks to be a more difficult problem and may require another scheduling layer above this, which would be analogous to the cognitive linking in the first model. The procedural logic requires verbs for understanding it, but that is a labelling problem. If the label is removed, we still have the relations between the node paths and so if at some level, the brain knows what each type is, it can relate that to the path through this architecture. Comparing the two models shows that the knowledge-to-experience link sets can match nodes at boundaries and also help to transition from ensemble to type. The logic level now looks relatively primitive, almost like a bee dance.

### 5.2 Realising Patterns as Self-Organising Units

To be consistent with the unit structure, patterns need to be realised as linked sets of them. There are probably many different ways to represent patterns and it would be analogous to the columnar structure that makes up the cortex region [22][17], for example. While in the neocortex, the columnar structure is a physical entity, maybe in other brain regions, the same type of functionality can be realised, even if the physical space is slightly different. Therefore, to realise the self-organising units, it is necessary to provide limiting constraints on the firing patterns. If an ensemble feeds a converging tree or hierarchy, which then feeds back into the ensemble, that loop is then a limiting constraint. The unit would also be more akin with the knowledge-based patterns and then expanding experience-based sets would be able to link these units together.

### 5.3 Linking Global Regions

If the structures prefer to converge for economy of energy [13], it then remains to make clear why they would diverge as well and the reason for this is to add new knowledge to the network. It has been shown in statistical physics [26], for example, that global properties in a network can be orderly, even if there is randomness at the local level. Even if individual units can behave slightly stochastically, there may still be global trends that would allow the larger network to behave in a more uniform manner. There is therefore also coherence in the network at the global level, above what each individual unit produces, allowing statistics to take over from intelligence. If brain regions started to fire at the same time for example, that would encourage growth in pattern sets at a global level and would encourage those sets to merge into a larger ensemble. So that would help to expand the network, again simply through Hebbian linking.

## 6   Conclusions

This paper has integrated the cognitive model further and shown it to be mathematically consistent. The reason for the ensembles and different types of hierarchy is clear and also the reason for transposing between them, where some hierarchies are the interface between different regions. Self-organising units of work can help with aggregation and both knowledge-based converging and experience-based diverging structures have a specific purpose. In fact, it is proposed that knowledge transpositions are what make the network flexible. This paper is different to the earlier research with the geometrically progressive design. Units of work converge from the ensemble to sets of abstract concepts and from them to types. Concept Trees may then not be about linking individual nodes, but about linking the self-organising units. They would grow by linking knowledge constructs together and through global network properties. A process analogous to stigmergy can be re-introduced and would take place in the memory region, where in a real brain it would be between the base and the nervous system, for example. It would help with pattern organisation and optimise their placement, so that related patterns can be found more easily. This is a nice idea, because the role of stigmergy in insect colonies is to optimise paths of the preferred routes, where the actor in question has no influence. The human brain has then evolved to add more intelligent functions on-top of that.

The two cognitive models can now be modelled as similar systems that differ in functionality only. This is particularly clear if the procedural logic model includes a more constructive scheduling layer on-top of its more primitive logical level. The lower memory structures can be the same and the upper 'intelligent' level can either store constraints on behaviour actions, or the behaviours themselves. Noted that spatial information should also be stored somewhere. The original designs are still very relevant, where both unique and shared pattern sets exist. The unique sets can be converted into representing functions that determine what the brain subsequently understands. The author posted a number of statements in an earlier paper [15], about requirements for building a brain-like model. They can probably be answered by using this model.

DCS 16 May 2021

## Appendix A

This appendix lists the two detailed cognitive models. Figure 5 shows the original model with the more basic upper level, while the preferred Figure 6 shows the second model with the behaviour upper level and more geometric structure.

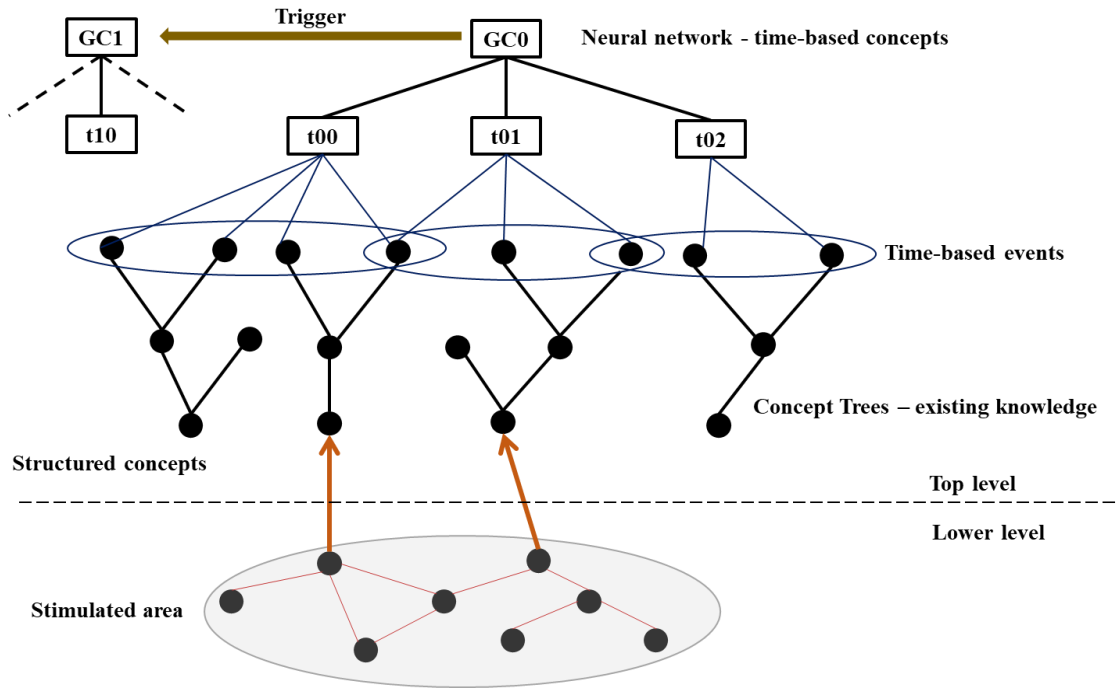

Figure 5. First detailed cognitive model from [12].

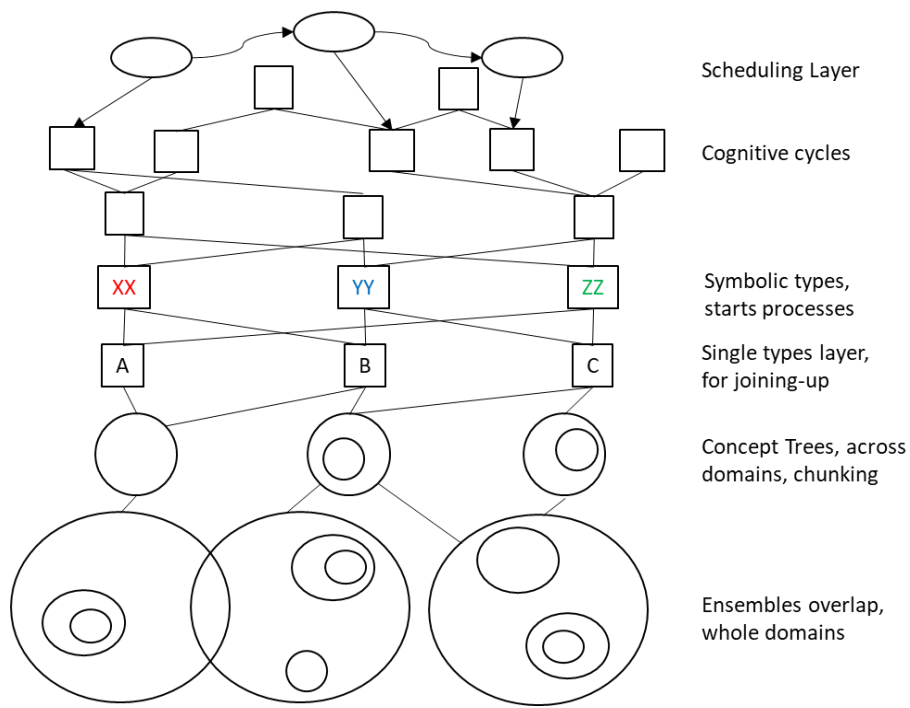

Figure 6. Second detailed cognitive model from [7].